\pdfoutput=1

\documentclass[11pt]{article}

\usepackage{ACL2023}

\usepackage{times}
\usepackage{latexsym}

\usepackage[T1]{fontenc}

\usepackage[utf8]{inputenc}

\usepackage{microtype}

\usepackage{inconsolata}

\usepackage{booktabs}
\usepackage{multirow}
\usepackage{graphicx}
\usepackage{makecell}
\usepackage{xcolor}
\usepackage{soul}

\definecolor{Seashell}{RGB}{255, 245, 238}
\definecolor{Firebrick4}{RGB}{255, 0, 0}
\definecolor{DarkOrange}{RGB}{255, 140, 0}
\definecolor{AntiqueWhite1}{RGB}{255, 239, 219}
\definecolor{DodgerBlue}{RGB}{30, 144, 255}
\definecolor{LightCyan}{RGB}{224, 255, 255}

\newcommand{\wrong}[1]{
  \begingroup
  \sethlcolor{Seashell}
  \textcolor{Firebrick4}{\hl{#1}}
  \endgroup
}

\newcommand{\key}[1]{
  \begingroup
  \sethlcolor{LightCyan}
  \textcolor{DodgerBlue}{\hl{#1}}
  \endgroup
}

\title{Multimodal Data Augmentation for Image Captioning using Diffusion Models}

\author{Changrong Xiao$^1$, \quad Sean Xin Xu$^1$, \quad Kunpeng Zhang$^{2}$ \\
  $^1$School of Economics and Management, Tsinghua University \\
  $^{2}$Department of Decision, Operations \& Information Technologies, University of Maryland \\
  \texttt{xcr21@mails.tsinghua.edu.cn} \\
  \texttt{xuxin@sem.tsinghua.edu.cn} \\
  \texttt{kpzhang@umd.edu} \\}

\begin{document}
\maketitle

\begin{abstract}
Image captioning, an important vision-language task, often requires a tremendous number of finely labeled image-caption pairs for learning the underlying alignment between images and texts. In this paper, we proposed a multimodal data augmentation method, leveraging a recent text-to-image model called \textit{Stable Diffusion}, to expand the training set via high-quality generation of image-caption pairs. Extensive experiments on the MS COCO dataset demonstrate the advantages of our approach over several benchmark methods, and particularly a significant boost when having fewer training instances. In addition, models trained on our augmented datasets also outperform prior unpaired image captioning methods by a large margin. Finally, further improvement regarding the training efficiency and  effectiveness can be obtained after intentionally filtering the generated data based on quality assessment.  
\end{abstract}

\section{Introduction}


Image captioning aims to automatically generate textual descriptions of visual content in an image, an important task at the intersection of natural language processing (NLP) and computer vision (CV) \citep{staniute2019systematic, atliha2020text, cornia2020meshed, turkerud2021image}. While impressive results have been achieved especially with emerging deep learning algorithms, most studies have been still tied to model optimization, extracting informative features, or developing better training techniques. Data, as a critical dimension that can significantly affect model performance, is greatly under-explored, despite a recent increase in interest \citep{feng-etal-2021-survey, turkerud2021image}.

Having large amounts of finely labeled image-text pairs in supervised image captioning tasks is often desired \citep{hendricks2016deep, zhu2022unpaired}. Existing image captioning datasets, such as MS COCO \citep{lin2014microsoft}, require human annotators to label images with descriptive sentences, which is laborious and time-consuming. Moreover, the collected images and annotated captions are possibly incomplete and lack variety, which limits the generalization ability of models trained on such datasets \citep{zhu2022unpaired}. 

To address the data issue, some researchers have attempted to leverage unpaired images and text, since they are easily obtained separately \citep{hossain2019comprehensive, kim2019image, laina2019towards, zhu2022unpaired}. This task is referred to as Unpaired Image Captioning (UIC), which has attracted considerable attention recently. Others have leveraged data augmentation to increase training data and improve model performance, including augmentation for images \citep{wang2018image, katiyar2021image} and textual captions \citep{atliha2020text, turkerud2021image, atliha2022text}. 

In this work, we attempt a more challenging situation where only textual data and no images are provided. We apply \textit{Stable Diffusion} \citep{rombach2022}, one of recent well-developed text-to-image models, to generate high-quality images with given text inputs. We further create several synthetic datasets to expand image-caption pairs based on text and/or image augmentation. Trained on these synthetic datasets, the image captioning models can be significantly improved over state-of-the-art UIC methods. We also extensively demonstrate that models trained on our multimodal-augmented datasets outperform previous image captioning data augmentation methods, especially when limited ground-truth data are provided. To further improve the performance, we assess and filter the augmented data based on quality measures. Overall, this paper makes the following contributions:

\begin{itemize}
    \item We develop and implement a multimodal data augmentation method to improve the performance of image captioning models. To our knowledge, this is among the first attempt to apply augmentation for both images and texts simultaneously in image captioning tasks. 
    \item We conduct extensive experiments and the results show that high-quality synthetic data generated by large pre-trained text-to-image models can significantly improve the quality of image captions in zero- and few-shot scenarios. Our study also provides a successful application of the \textit{Stable Diffusion} model. 
    \item Finally, we intend to release the code and the augmented datasets to the community\footnote{\url{https://github.com/Xiaochr/Multimodal-Augmentation-Image-Captioning}}. With our effective data generation and quality assessment for image-caption pairs, we demonstrate that training datasets can be constructed without expensive human annotation efforts for supervised vision-language tasks. 
    
\end{itemize}

\section{Related Work}

\subsection{Data Issue in Image Captioning}

In recent years, image captioning models have developed rapidly benefited from deep learning algorithms. Encoder-Decoder architecture is one of the most common and effective model frameworks, with Convolutional Neural Networks (CNN) encoders to obtain image features and Recurrent Neural Networks (RNN) for decoding them into natural language \citep{hossain2019comprehensive}. Attention mechanisms \citep{xu2015show, lu2017knowing, anderson2018bottom, huang2019attention} and Transformers \citep{li2019entangled, cornia2020meshed} are actively used and significantly boosting performance. 

Training image captioning models with fully supervised methods requires tremendous paired image-caption data. With a lack of training data, even state-of-the-art models rarely perform well. To tackle this issue of limited labeled data, unpaired image captioning (UIC) and data augmentation approaches have been proposed by researchers. 

\paragraph{Unpaired Image Captioning}


The UIC task aims to generate captions from models trained using unpaired images and captions, which are easily obtained separately from various sources (e.g., the web). This has attracted significant attention from researchers given the high cost of obtaining paired images and texts. For example, \citet{gu2018unpaired} implemented language pivoting with extra Chinese caption information. \citet{feng2019unsupervised} proposed an unpaired image captioning framework by learning a visual concept detector. \citet{laina2019towards} exploited large text corpora outside the dataset and learned shared multi-modal embeddings for images and sentences. More recently, \citet{zhu2022unpaired} tackled the visual concept recognition stage for UIC aided by only image-level class labels. On the other hand, semi-supervised learning methods also come to assist. \citet{chen2016semi} generates missing visual information based on textual data. \citet{kim2019image} implemented GANs to assign pseudo-labels to unlabeled images. Most UIC studies are in common that they exploit beyond image captioning data and require more or less some auxiliary information or expensive annotations. 

\paragraph{Data Augmentation}
is to create additional training data with diversity. They have been widely applied to images and texts separately in various machine learning tasks. However, augmentation is rarely used in vision-language tasks, such as image captioning \citep{atliha2020text}. 

Most data augmentation in image captioning studies focus on either image or caption, rather than both simultaneously. \citet{wang2018image, katiyar2021image} used standard image transformations like cropping, flipping, and mirroring to manipulate image data. However, these image augmentations may introduce noises when transformations distort the semantics of images. For caption augmentation, \citet{atliha2020text} applied synonym replacement and paraphrasing sentences using BERT \citep{devlin2018bert}. Word permutation/replacement \citep{cui2018learning}, back translation \citep{turkerud2021image}, and other NLP augmentation techniques were also used in prior literature. 

However, multi-modal augmentation, where both images and texts are modified at the same time, is a relatively underexplored direction \citep{hartmann2022interactive}. \citet{feng2021survey} introduced an approach to combine CutMix \citep{yun2019cutmix} and caption editing, by inserting patches cut out from a different image and modifying the caption such that it correctly describes the new image. But the authors did not implement this method and no experimental results were provided.

\subsection{Text-to-Image Synthesis}

Text-to-Image synthesis is a challenging multimodal task of generating a high-quality image conditioned on a descriptive text \citep{du2022survey, yang2022diffusion}. This field had been dominated by Generative Adversarial Networks (GANs) \citep{goodfellow2014gan}, Variational Autoencoders (VAEs) \citep{kingma2013auto}, and other generative models. Nonetheless, these models suffer from some drawbacks. For example, GANs are difficult to optimize \citep{mescheder2018training} and are confined to data with limited variability \citep{brock2018large, karras2019style}; VAEs are more efficient to generate high-resolution images, but the sample quality is not good enough \citep{rombach2022}. 

A recently emerged family of deep generative models, diffusion models \citep{Ho2020ddpm, song2021scorebased}, has shown their impressive power and beats GANs in text-to-image synthesis \citep{dhariwal2021diffusion}. While state-of-the-art models like \textit{DALLE-2} \citep{ramesh2022hierarchical} and \textit{Imagen} \citep{saharia2022photorealistic} are able to generate images of surprisingly high quality, their inference is expensive and time-consuming \citep{rombach2022}. In this study, we adopt an improved Latent Diffusion Model, \textit{Stable Diffusion} \citep{rombach2022}, as our text-to-image component. \textit{Stable Diffusion} can generate high-resolution images comparable to state-of-the-art models with affordable computational resources and times. 

In image captioning tasks, generative models are rarely applied. \citet{kim2019image} used \textit{CycleGAN} \citep{zhu2017unpaired} as a baseline for their unpaired image captioning, which proved to be less effective. In the most recent work, \citet{li2022dall} demonstrated that the best caption for an image is the one that leads to the best reconstruction of the original image, using \textit{Stable Diffusion} as the text-to-image model and \textit{Flamingo} \citep{alayrac2022flamingo} as the inverse one. However, diffusion models have not been utilized to improve the quality of generated image captions, despite the impressive text-to-image results they have achieved. 


\section{Method}

Our proposed image captioning system consists of two parts. In the first part, we implement our image-caption pair generation to construct the synthetic dataset. The dataset can be further expanded via text or image augmentation methods. The second part shows how we train and evaluate two selected image captioning models based on the constructed data in part one. 

\subsection{Multimodal Augmentation} 

We use MS COCO \citep{lin2014microsoft} as the base dataset to perform augmentation. MS COCO is a large and commonly used image captioning dataset, with 123,287 images and 616,767 captions in total. 

In our multimodal augmentation, we first apply \textit{Stable Diffusion} to generate one image for each COCO caption while discarding images with NSFW content. Then we pair the generated image and the true COCO caption to form a base synthetic dataset (denoted by $SD_{base}$). 

\begin{figure*}[h]
    \centering
    \includegraphics[width=1\textwidth]{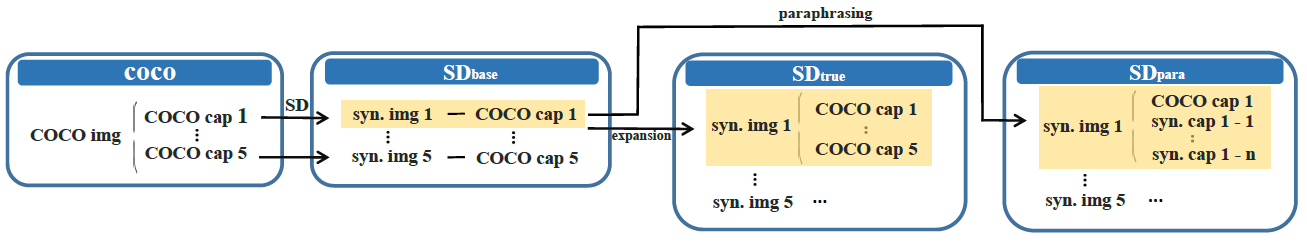}
    \caption{An illustrative example of how to construct the three synthetic datasets ($SD_{base}$, $SD_{true}$, and $SD_{para}$) using text-to-image generation and caption expansion. \textit{COCO img/cap} represents images/captions in the original COCO dataset, \textit{syn. img/cap} represents synthetic images/captions. }
    \label{datasets}
\end{figure*}

Prior research has found that image captioning models trained on images with more diverse descriptions can achieve better performance \citep{devlin2018bert, atliha2020text}. Thus, one caption for one image in $SD_{base}$ may not be enough, and expanding the caption for each generated image is desirable. In this study, we expand captions using two strategies: (i) from ground-truth COCO captions, (ii) from automatic caption generation via text augmentation (e.g., paraphrasing). The obtained datasets are denoted as $SD_{true}$ and $SD_{para}$, respectively. Figure \ref{datasets} illustrates the way to construct three augmented datasets through one example in COCO data (i.e., one example indicates one image and its 5 corresponding captions). 

\begin{itemize}
    \item $SD_{base}$ consists of 5 image-caption pairs, one generated image per true caption: (synthetic image 1, true caption 1), $\cdots$, (synthetic image 5, true caption 5).
    \item $SD_{true}$ expands $SD_{base}$ based on the assumption that all 5 captions are relevant. Thus, each generated image from one caption is mapped to all 4 other captions and itself: (synthetic image 1 from true caption 1, true caption 1), $\cdots$, (synthetic image 1 from true caption 1, true caption 5), $\cdots$, (synthetic image 2 from true caption 2, true caption 1), $\cdots$, (synthetic image 2 from true caption 2, true caption 5), $\cdots$, (synthetic image 5 from true caption 5, true caption 1), $\cdots$, (synthetic image 5 from true caption 5, true caption 5). Thus, $SD_{true}$ is expanded to $25$ image-caption pairs for just one example.
    \item $SD_{para}$ is built upon $SD_{base}$. Each true caption is augmented via $k$ times of paraphrasing. Thus, $SD_{para}$ has: (synthetic image 1, $k$ synthetic captions), $\cdots$, (synthetic image 5, $k$ synthetic captions). Note that we discard those generated paraphrases that are exactly the same. Therefore, the amount of corresponding captions for each generated image in $SD_{para}$ is $k+1$ (i.e., one true caption and $k\leq 5$ augmented ones).
\end{itemize}   


Next, we create several additional datasets while accounting for the scenario where the availability of labeled data (i.e., pairs) varies. Specifically, one typical case would be that only a small subset of COCO image-caption pairs while all captions are available. In this case, our proposed multimodal augmentation can be applied to increase pairs. For example, we first keep these provided image-caption pairs. For the rest captions, we apply augmentation as in $SD_{base}$ to create more image-caption pairs. Such a new dataset is denoted as $n\% \  COCO + SD_{base}$, where $n$ is the percentage of the original pairs in COCO. For example, $10\% \  COCO + SD_{base}$ means $10\%$ true pairs in COCO combined with $SD_{base}$ from the rest $90\%$ captions. Similarly, $n\% \  COCO + SD_{true}$ and $n\% \  COCO + SD_{para}$ can be obtained. 

In contrast to our multimodal augmentation method, there exist many uni-modal methods. For example, paraphrasing is a widely adopted text augmentation via a large language model similar to \citet{atliha2020text}. For image augmentation, prior studies use random flipping combined with random perspective transformation \citep{wang2018image, katiyar2021image}. Based on these, we create two more datasets. The generated paraphrases are added into the COCO dataset as additional captions, denoted as $COCO_{text}$. We replace the original COCO images with the augmented images to obtain $COCO_{image}$. 

\subsection{Selected Image Captioning Models}

Now we turn towards introducing two selected image captioning models upon which we evaluate the effectiveness of the augmented datasets. Our objective is to investigate whether experiment results are model-specific and whether our constructed datasets can be used to improve image captioning models in general. To do so, we fix the image-captioning model to  FC model, which is relatively simple and small-sized. We also choose another more advanced Transformer-based model for robustness. 

\paragraph{FC Model} is a frequently used model in many image captioning studies \citep{vinyals2015show, karpathy2015deep, rennie2017self, luo2018discriminability}. In the FC Model, a CNN architecture is first embedded to extract visual features for each image. Then the extracted feature embeddings are processed by Long Short-Term Memory (LSTM) modules \citep{hochreiter1997long} to generate captions. 

\paragraph{Transformer-based Model} Recently, Transformer \citep{vaswani2017attention} models are boosting the performance of various deep learning tasks, including image captioning \citep{cornia2020meshed, luo2021dual, zhou2022compact}. In this study, we do not choose a specific Transformer-based image captioning model but rather a generic architecture: it first extracts visual features using a bottom-up approach \citep{anderson2018bottom}, and a basic Transformer module with self-attention is applied to decode the visual features to textual captions.

\section{Experiments and Results}

\subsection{Experiment Setup}

In this study, we apply the \textit{Stable Diffusion} model version 1-4\footnote{\url{https://huggingface.co/CompVis/stable-diffusion-v1-4}} implemented in \textit{huggingface} to generate synthetic images with given captions. We keep default settings. Model parameters are frozen during the generation process. The NSFW images are automatically detected, and we do the image generation again for these images until they are suitable to be included in our synthetic datasets. Finally, $566,747$ synthetic images are generated. 

Synthetic caption expansion for $SD_{para}$ and text augmentation for $COCO_{text}$ use the same paraphrasing approach based on true COCO captions: A pre-trained T5 model \citep{raffel2020exploring} for paraphrasing\footnote{\url{https://huggingface.co/Vamsi/T5_Paraphrase_Paws}}. For image augmentation, we apply $RandomHorizontalFlip$ and $RandomPerspective$ transformation functions implemented by \textit{torchvision}, with both transformation probabilities set to $0.5$. 

The implementation of feature extraction, model training, and model evaluation are mainly adapted from \citet{luo2018discriminability}\footnote{\url{https://github.com/ruotianluo/ImageCaptioning.pytorch}}. We train image-captioning models based on Karpathy's split \citep{karpathy2015deep} of the training set with the fully supervised method and a more effective CIDEr score optimization \citep{luo2018discriminability}. An early-stop strategy is also adopted, with maximum training epochs of $30$ and $15$ for the FC model and the Transformer-based model, respectively. Regarding the evaluation, we use standard metrics for image captioning tasks, including BLEU \citep{papineni2002bleu}, METEOR \citep{banerjee2005meteor}, ROUGE \citep{lin2004rouge}, CIDEr \citep{vedantam2015cider} and SPICE \citep{anderson2016spice}. All evaluation metrics remain the same as \citet{luo2018discriminability}, and all the evaluations of COCO dataset and synthetic datasets are on Karpathy's test split. We set the seed as $42$ to run all experiments with 2 RTX A4000 16G GPUs. In the sequel, we show three experiments and their results to demonstrate the effectiveness of our multimodal data augmentation upon which image captioning tasks are performed and evaluated.


\subsection{Experiment 1: Fully Synthetic Dataset}

In this experiment, we train models using 3 datasets where all images are synthetic: $SD_{base}$, $SD_{para}$, and $SD_{true}$. We compare the performance of image captioning models (i.e., FC and Transformer) with state-of-the-art UIC methods. Note that UIC task assumes unpaired but true images were available, which is a more informed condition than our scenario where no true images are available. The experiment results are shown in Table \ref{exp1}. Since the results of the two models are fairly consistent, we show the results from the FC model here and include those from the Transformer-based model in Appendix. 

\begin{table}[h]
\centering
\small
\begin{tabular}{@{}p{0.32\columnwidth}|p{0.08\columnwidth}p{0.08\columnwidth}p{0.08\columnwidth}p{0.08\columnwidth}p{0.08\columnwidth}@{}}
\toprule
                           \textbf{Training Set} & \textbf{B4} & \textbf{M} & \textbf{R} & \textbf{C} & \textbf{S} \\ \midrule
Fully paired COCO  & 28.7   & 24.4   & 52.4  & 92.4  & 17.6  \\ \midrule
\citet{gu2018unpaired}                    & 5.4    & 13.2   & -     & 17.7  & -     \\
\citet{feng2019unsupervised}                      & 18.6   & 17.9   & 43.1  & 54.9  & 11.1  \\
\citet{laina2019towards}                      & 19.3   & 20.2   & 45.0  & 61.8  & 12.9  \\
\citet{zhu2022unpaired}                     & 21.5   & 20.1   & 45.8  & 65.7  & 13.6  \\ \midrule
Our $SD_{base}$        & 23.6   & 21.7   & 48.6  & 75.4  & 15.3  \\
Our $SD_{para}$   & 23.4   & 21.6   & 48.9  & 75.5  & 15.4    \\
Our $SD_{true}$        & \textbf{25.6}   & \textbf{22.6}   & \textbf{50.1}  &  \textbf{81.0}  & \textbf{15.6}   \\ \bottomrule
\end{tabular}
\caption{Performance comparison between our multimodal augmentation and UIC. The first row shows the performance of FC model trained on the original COCO dataset, which is regarded as a comparison benchmark; the middle rows list the performances of 4 commonly used UIC baselines reported in prior studies; and the bottom rows are the performances of FC model trained on our 3 synthetic datasets. Note that "B4", "M", "R", "C", and "S" stand for "BLEU-4", "METEOR", "ROUGE", "CIDEr", and "SPICE", respectively. }
\label{exp1}
\end{table}

We have the following observations: (1) training with our basic multimodal-augmented dataset ($SD_{base}$) outperforms UIC baselines in all metrics by a large margin. In other words, the pre-trained text-to-image model can be used to generate images based on given captions to train image captioning models. Image captioning models can be trained in a simple fully supervised manner on synthetic image-caption pairs, instead of semi-/un-supervised way based on true but unpaired image-caption data in UIC tasks. Even without true images, the synthetic data performs surprisingly well in the downstream captioning task, which shows great potential for the application of synthetic data generation. (2) Expanding captions with synthetic paraphrases ($SD_{para}$) yields better results. Trained on $SD_{true}$, a significant performance improvement is further achieved. This may be due to that text obtained by paraphrasing is less diverse than real captions annotated by humans. This is also in line with previous studies \citep{atliha2020text, turkerud2021image}, where researchers found that better model performance would be achieved if images in the training set have more captions with higher diversity.

\subsection{Experiment 2: Few-shot Learning}

\begin{figure*}[h]
    \centering
    \includegraphics[width=0.45\textwidth]{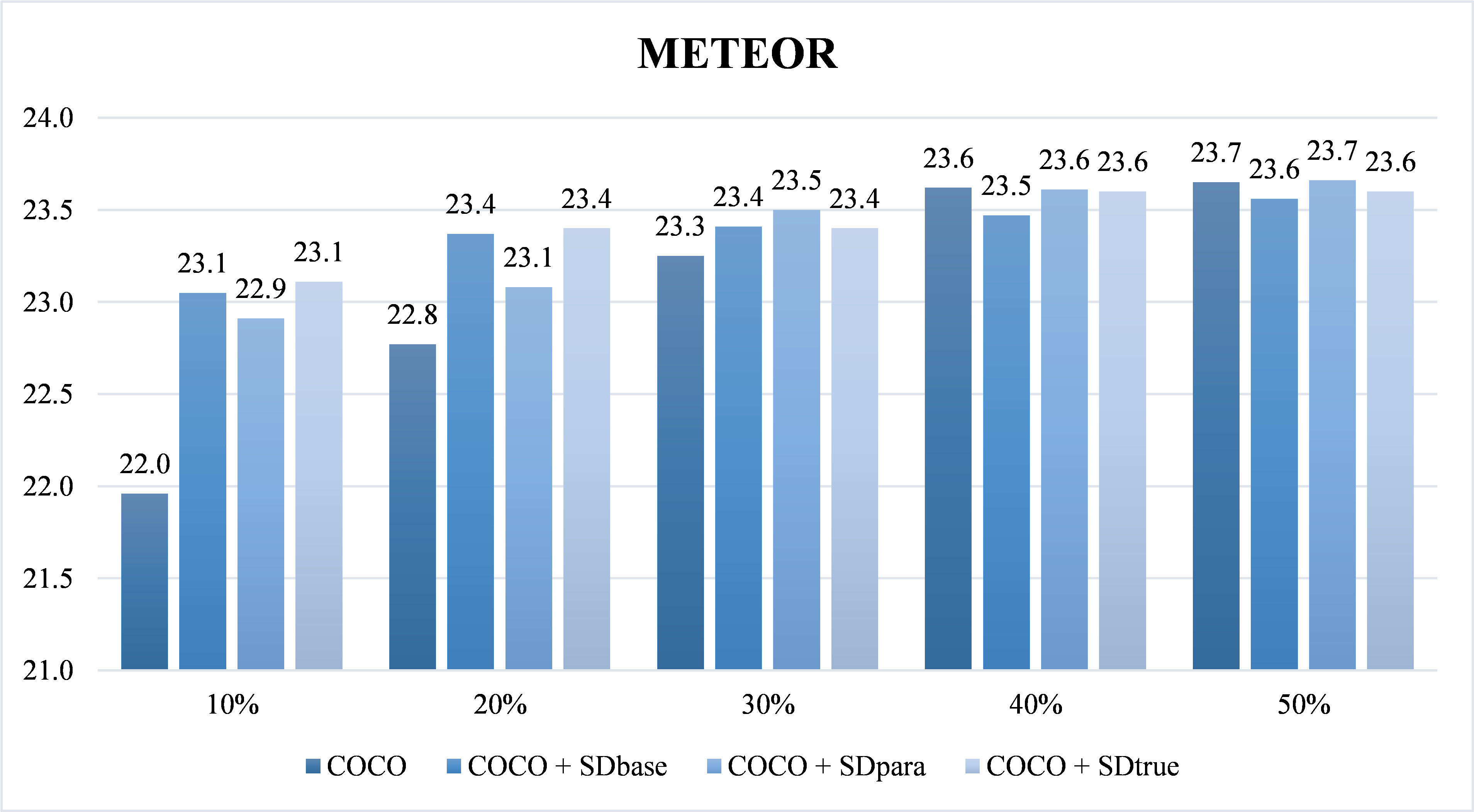}
    \includegraphics[width=0.45\textwidth]{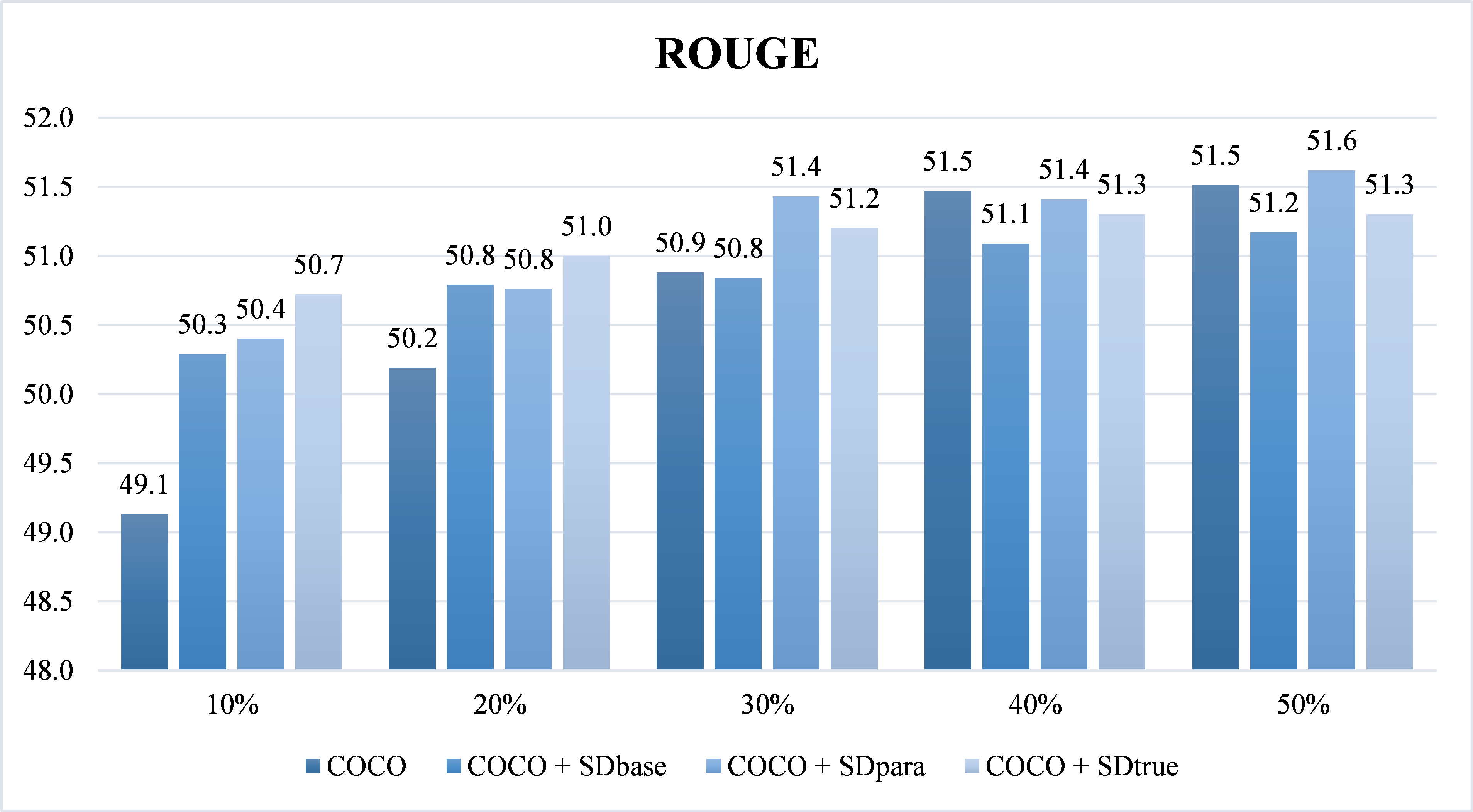}
    \includegraphics[width=0.45\textwidth]{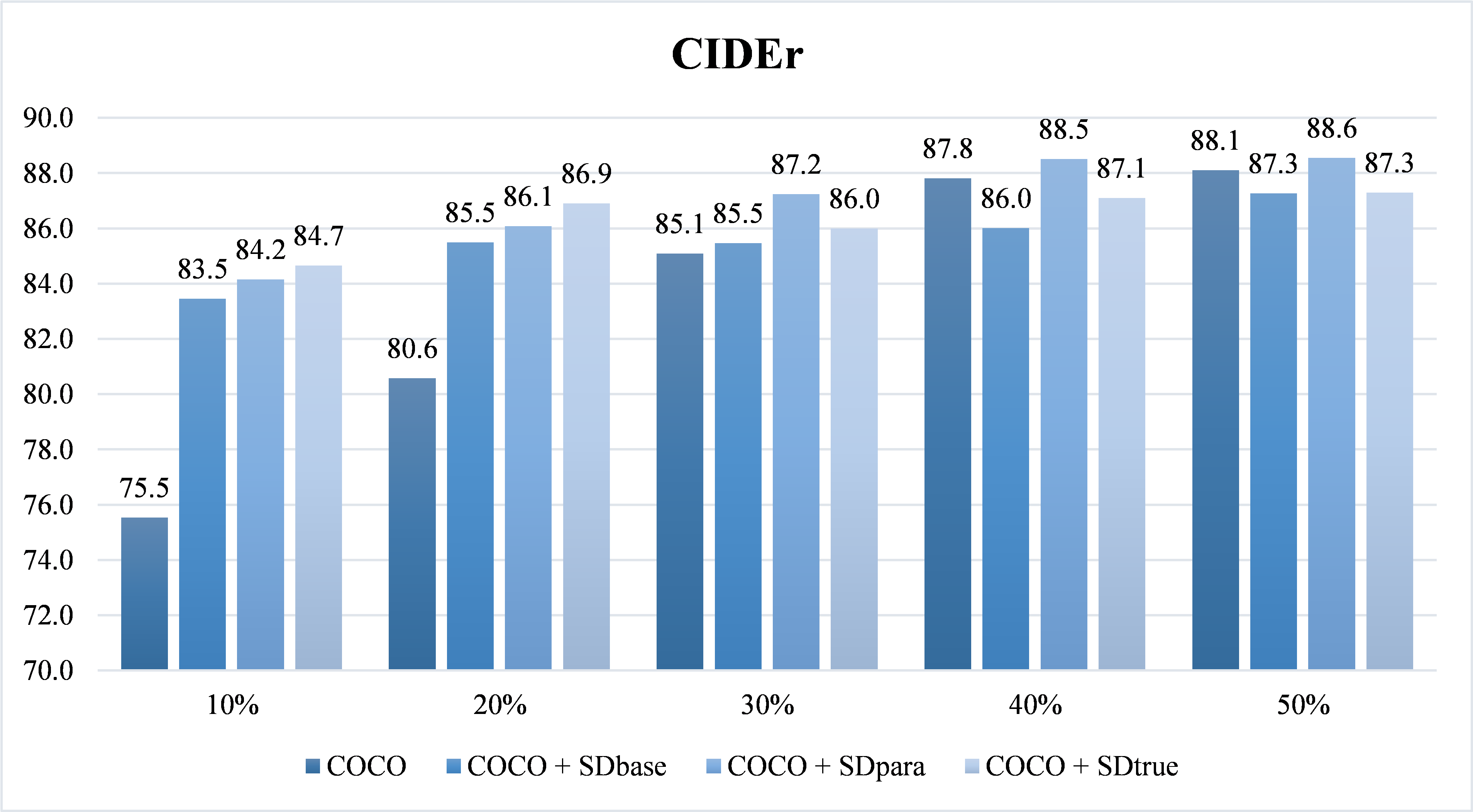}
    \includegraphics[width=0.45\textwidth]{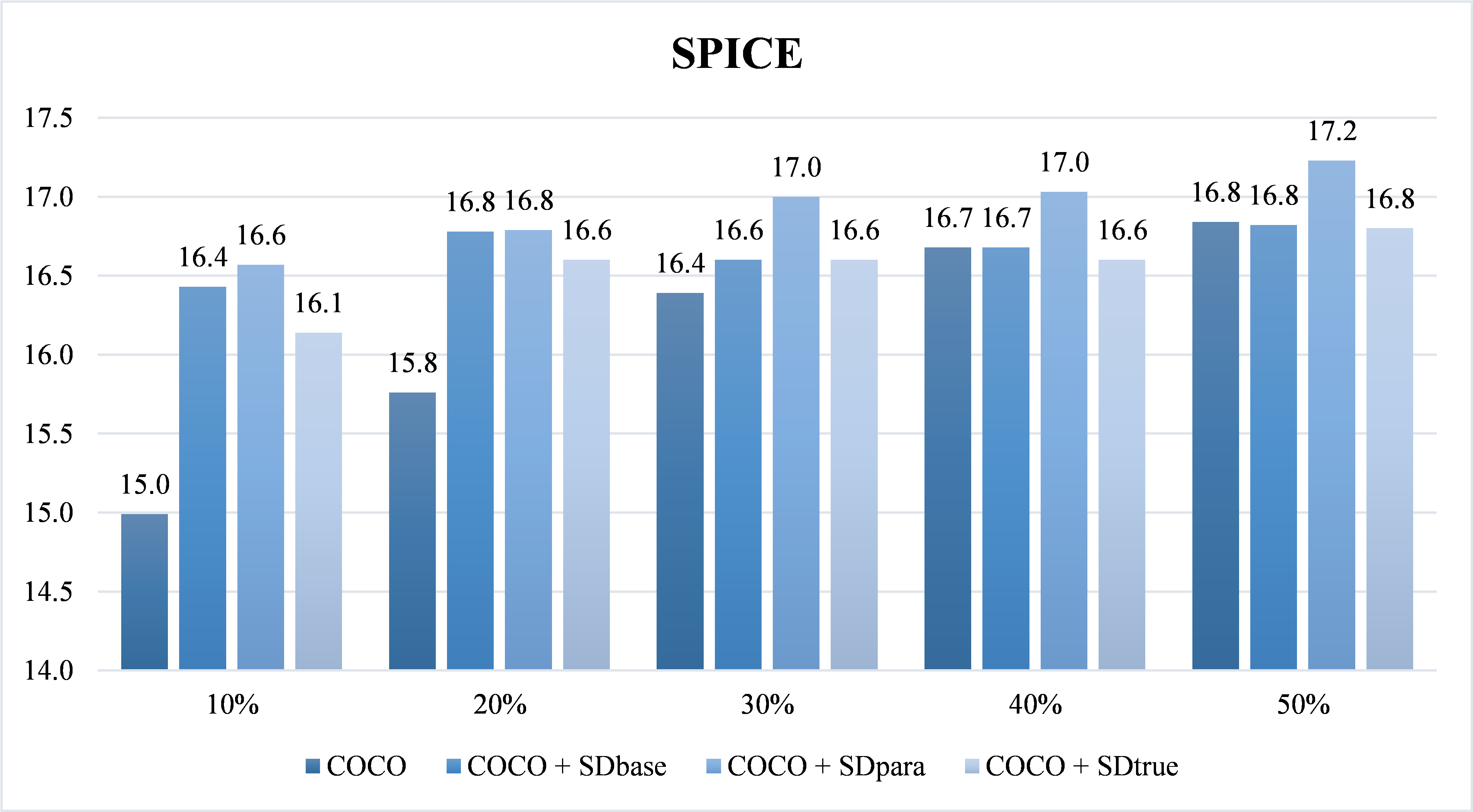}
    \caption{Image captioning performance of three multimodal augmented datasets with different portions of COCO data. The five groups are situations with $10\%$ to $50\%$ COCO data, respectively. Within each group, only $n\%$ COCO, $n\%$ COCO with $SD_{base}$, $n\%$ COCO with $SD_{para}$, and $n\%$ COCO with $SD_{true}$ are presented from left to right. The figure shows only part of the evaluation metrics. For more details, see Table \ref{exp2-fc-full} in Appendix. }
    \label{exp2-fig}
\end{figure*}

Next, we intend to understand the performance of our multimodal data augmentation method when the ground-truth data is limited. To do so, we first sample the original COCO data with different percentages, ranging from $10\%$ to $50\%$, and then apply text, image, and multimodal data augmentation methods to them. In addition, we assume that all captions in  COCO are available, so that multimodal data augmentation is to complete the unavailable images. This assumption is reasonable since the textual information alone is relatively easy to obtain. 

\begin{table}[ht]
\centering
\small
\begin{tabular}{@{}l|ccccc@{}}
\toprule
\textbf{Training Set}        & \textbf{B4} & \textbf{M} & \textbf{R} & \textbf{C} & \textbf{S} \\ \midrule
10\% COCO           &  24.3   & 22.0   & 49.1  & 75.5  & 15.0  \\ \midrule
10\% $COCO_{text}$    &  23.8   & 21.6   & 49.1  & 74.6  & 15.0  \\
10\% $COCO_{img}$     &  24.1   & 22.0   & 49.0  & 75.7  & 15.0  \\ \midrule
10\% COCO + $SD_{base}$ &  25.8   & 23.1   & 50.3  & 83.5  & 16.4  \\
10\% COCO + $SD_{para}$ &  26.0   & 22.9   & 50.4  & 84.2  & \textbf{16.6}  \\
10\% COCO + $SD_{true}$ &  \textbf{26.2}   & \textbf{23.1}   & \textbf{50.7}  & \textbf{84.7}  & 16.1  \\ \bottomrule
\end{tabular}
\caption{Performance of the FC model when 10\% of pairs in COCO is available. }
\label{exp2-fc-10}
\end{table}

From the results (shown in Table \ref{exp2-fc-10}), we find that when the amount of training instances is very limited ($10\%$), basic data augmentations for uni-modality (images or text) have nearly no effect on the performance of image captioning (i.e., comparing rows 2 and 3 with row 1). By contrast, our multimodal data augmentation can significantly improve the performance (i.e., comparing the last 3 rows with row 1). Moreover, caption expansion for synthetic images can further improve performance (i.e., comparing row 5: $10\%+SD_{para}$ with row 4: $10\%+SD_{base}$), which is consistent with findings in Experiment 1. 

All experiment results from $10\%$ to $50\%$ COCO data are summarized in Figure \ref{exp2-fig}. As the size of ground-truth data increases, the gain of performance improvement from data augmentation gradually decreases. When the true data reaches $40\%$, the results of augmenting with $SD_{base}$ become worse than those with only $40\%$ true data. One plausible explanation is that $40\%$ true data in COCO is large enough for training a decent model. Adding synthetic images might bring more noise which could hurt the performance. 

Among the three multimodal augmentation methods, true caption expansion of synthetic images performs better than automatic paraphrasing expansion when the true data is very limited ($10\%$). When the amount of true data increases ($\geq 40\%$), however, true caption expansion cannot achieve the same results as $n\%$ COCO baseline. Nonetheless, automatic paraphrasing can still improve the model performance in this case. This may be due to novel words outside the COCO vocabulary, which is introduced by the paraphrasing model.

\subsection{Experiment 3: Synthetic Data Filtering}

In this section, we attempt to further improve the performance with augmented datasets in the image captioning task. 

We noticed that images generated by \textit{Stable Diffusion} possibly mismatch textual  descriptions, e.g., omitting important objects or having unnatural distortions. Selecting high-quality images that align text well is highly desired. Previous studies have come up with various methods to do this. For example, \citep{salimans2016improved} introduced Inception Score (IS) to evaluate GAN-generated images. \citep{heusel2017gans} presented Frechet Inception Distance (FID) as the golden standards to measure image quality. These metrics may not be directly applied in our context. Part of the reason is that they measure either the distortion of generated images to real ones, or the difference between two distributions, while our objective aims to find images that better match text. 



In addition, the evaluation should not depend on the real reference image and text. Therefore, we consider the following three aspects when assessing image quality in our context:
\begin{itemize}
    \item The quality of the image itself, measured by one of the recent proposed no-reference image quality assessment (NR-IQA) metrics, called \textit{MUSIQ} \citep{ke2021musiq}. 
    \item The similarity between the synthetic image and the corresponding input text. \textit{CLIPScore} \citep{hessel2021clipscore} is a reference-free metric for measuring captioning performance based on CLIP embeddings, which can be used to measure the coherence of generated image-caption pairs. 
    \item Whether the synthetic image reflects important objects described in the caption. \textit{VIFIDEL} \citep{madhyastha2019vifidel} is a newly proposed quality measurement method for image captioning tasks, using object detection models to recognize main objects in images, and then calculating the similarity between the objects and the description text.
\end{itemize}

We use \textit{MUSIQ}, \textit{CLIPScore}, and \textit{VIFIDEL} as our data filtering criteria, and find that \textit{CLIPScore} is the most effective metric among the three. Selecting top $50\%$ of data with the highest \textit{CLIPScore} could achieve similar results to using full synthetic datasets in Experiment 1 (see Table \ref{exp3-fc-select}). This indicates that it is unnecessary to train the model with full synthetic data. Data selection based on quality assessment can make the training more efficient without sacrificing performance. 

\begin{table}[h]
\centering
\small
\begin{tabular}{@{}l|ccccc@{}}
\toprule
\textbf{Training Set}    &  \textbf{B4} & \textbf{M} & \textbf{R} & \textbf{C} & \textbf{S} \\ \midrule
COCO             &  28.7   & 24.4   & 52.4  & 92.4  & 17.6  \\ \midrule
$SD_{base}$          &  \textbf{23.6}   & 21.7   & 48.6  & \textbf{75.4}  & 15.3  \\
Selected $SD_{base}$ &  23.5   & \textbf{22.2}   & \textbf{48.9}  & 75.3  & \textbf{15.6}  \\ \midrule
$SD_{para}$          &  \textbf{23.4}   & 21.6   & \textbf{48.9}  & \textbf{75.5}  & 15.4  \\
Selected $SD_{para}$ &  23.2   & \textbf{22.0}   & 48.7  & 75.0  & \textbf{15.6}  \\ \midrule
$SD_{true}$          &  \textbf{25.6}   & 22.6   & 50.1  & 81.0  & 15.6  \\
Selected $SD_{true}$ &  25.5   & \textbf{22.7}   & \textbf{50.2}  & \textbf{81.0}  & \textbf{15.7}  \\ \bottomrule
\end{tabular}
\caption{Performance of the FC model trained on three augmented datasets under data filtering with top $50\%$ \textit{CLIPScore}. }
\label{exp3-fc-select}
\end{table}

Since data selection is more effective on $SD_{true}$ among the three augmented datasets, we further explore the impact of data filtering in limited data situations with $SD_{true}$. We find in Table \ref{exp3-scarce} that data filtering actually improves the model performance. A significant boost still exists when the volume of true data gets larger, which surpasses the improvement by $SD_{para}$. This indicates that a suitable data filtering can improve both training efficiency and image captioning performance when true labeled data is limited. 

\begin{table}[h]
\centering
\small
\begin{tabular}{@{}p{0.38\columnwidth}|ccccc@{}}
\toprule
\textbf{Training Set}       & \textbf{B4}   & \textbf{M}    & \textbf{R}    & \textbf{C}    & \textbf{S}    \\ \midrule
10\% COCO + $SD_{para}$         & 26.0 & 22.9 & 50.4 & 84.2 & \textbf{16.6} \\
10\% COCO + $SD_{true}$          & 26.2 & 23.1 & 50.7 & 84.7 & 16.1 \\
10\% COCO + selected $SD_{true}$ & \textbf{26.8} & \textbf{23.3} & \textbf{51.1} & \textbf{86.2} & 16.5 \\ \midrule
20\% COCO + $SD_{para}$           & 26.3 & 23.1 & 50.8 & 86.1 & \textbf{16.8} \\
20\% COCO + $SD_{true}$          & 26.9 & 23.4 & 51.0 & 86.9 & 16.6 \\
20\% COCO + selected $SD_{true}$ & \textbf{27.1} & \textbf{23.5} & \textbf{51.2} & \textbf{87.2} & 16.7 \\ \midrule
30\% COCO + $SD_{para}$           & 27.0 & 23.5 & 51.4 & 87.2 & \textbf{17.0} \\
30\% COCO + $SD_{true}$          & 26.9 & 23.4 & 51.2 & 86.0 & 16.6 \\
30\% COCO + selected $SD_{true}$ & \textbf{27.0} & \textbf{23.7} & \textbf{51.4} & \textbf{87.8} & 16.9 \\ \midrule
40\% COCO + $SD_{para}$           & 27.3 & 23.6 & 51.4 & \textbf{88.5} & \textbf{17.0} \\
40\% COCO + $SD_{true}$          & 27.1 & 23.6 & 51.3 & 87.1 & 16.6 \\
40\% COCO + selected $SD_{true}$ & \textbf{27.3} & \textbf{23.7} & \textbf{51.4} & 88.3 & 17.0 \\ \midrule
50\% COCO + $SD_{para}$           & 27.5 & 23.7 & \textbf{51.6} & 88.6 & \textbf{17.2} \\
50\% COCO + $SD_{true}$          & 27.1 & 23.6 & 51.3 & 87.3 & 16.8 \\
50\% COCO + selected $SD_{true}$ & \textbf{27.6} & \textbf{23.7} & 51.5 & \textbf{89.3} & 16.9 \\ \bottomrule
\end{tabular}
\caption{The performance of FC model in limited-data settings with synthetic data selection based on the top $50\%$ \textit{CLIPScore} criterion. }
\label{exp3-scarce}
\end{table}

However, the improvement for $SD_{base}$ and $SD_{para}$ are not significant. And the image captioning performance even decreases in some cases for the Transformer-based model (see Table \ref{exp2-tfm-full} in Appendix). Therefore, \textit{CLIPScore} and the other two criteria are not golden standards to select high-quality data that are suitable for the captioning task, even though they have been testified to perform well in image-caption quality evaluation \citep{madhyastha2019vifidel, hessel2021clipscore}. 

A similar discrepancy between generated data quality and downstream task performance has been reported in a prior image classification task \citep{ravuri2019classification}. Authors found that although the GAN-generated images receive high scores close to those of true images, the classification model trained on fully synthetic images has a much lower accuracy than those trained on true images. Following this thread, we calculate the three metrics for both COCO data and synthetic data. We have a similar finding that three quality measures under the synthetic data are close level to those under true data (see Table \ref{exp3-iqa}). However, when completely replacing the true data with the synthetic data as the training (e.g., in Experiment 1), the performance is quite lower than that of the true data (i.e., CIDEr score: $81.0$ vs. $92.4$). In this way, we extend findings in \citet{ravuri2019classification}'s study to the image captioning task.

\begin{table}[h]
\centering
\small
\begin{tabular}
{@{}l|ccc@{}}
\toprule
                & \textbf{MUSIQ} & \textbf{CLIPScore} & \textbf{VIFIDEL} \\ \midrule
COCO data        &      69.8   & 78.4      &    35.8      \\
Synthetic data    & 69.6        & 80.1      & 34.4          \\ \bottomrule
\end{tabular}
\caption{Comparison of multimodal data quality assessment using true vs. augmented image-caption pairs. }
\label{exp3-iqa}
\end{table}

\section{Conclusion}

We developed a multimodal data augmentation method for the image captioning task, leveraging the power of diffusion models in image synthesis. It outperforms uni-modal methods on the MS COCO dataset for two typical image captioning models, especially when the amount of true labeled data is limited. It also performs significantly well comapred to UIC methods with fully synthetic images. 

Our study is an early attempt to combine image and text modals via two inverse processes: text-to-image and image-to-text. The effectiveness of synthetic multimodal data used as the training set was empirically verified, and better performance can be further achieved with data filtering. Though synthetic data are not able to replace true data, it is worth exploiting the potential of multimodal data synthesis and its applications in various downstream applications in the future.

\section*{Limitations}

Below we discuss three limitations of our multimodal data augmentation method. (1) The most noteworthy one is the quality of synthetic data, a common challenge seen in many data generation studies. We have explored some quality assessment metrics for image-caption pairs and tested their performance, specifically in the captioning task. Effective and generalized multimodal data quality assessment still remains an open question for future research, which we believe is a valuable direction. (2) The quality and flexibility of synthetic images are bounded by the ability of text-to-image models we apply. For example, if we intend to generate synthetic training datasets for the human face recognition task, good performance is unlikely to achieve since \textit{Stable Diffusion} is not good at drawing human faces. (3) Computational resource is needed to generate a large synthetic image dataset, even with lightweight models like \textit{Stable Diffusion}. In this study, generating the whole synthetic COCO images took us about 2 weeks with 2 RTX A4000 16G GPUs, which is demanding for practitioners and researchers. 

\section*{Ethics Statement}

In this study, \textit{Stable Diffusion}, pre-trained on enormous publicly available online images, is applied. However, some pointed out that several paintings created by online artists were included without authorization. Whether it harms the originality of artists or involves privacy issues is still under a heated debate. Developers of these large pre-trained text-to-image models have taken actions to eliminate any ethical concerns, for example, removing images without authorization from the training set. 

This may be an underlying problem for us, but our focus is the capability of these text-to-image models. Our method of generating higher-quality image-caption pairs can be applied in a wider range of domains (for example, as an assistant for those visually impaired in education), leading to greater impacts on society and human well-being.


\bibliography{anthology,custom}
\bibliographystyle{acl_natbib}

\appendix

\section{Synthetic Datasets}

Descriptive statistics of our augmented datasets are shown in Table \ref{data-stats}. 

\begin{table}[h]
\centering
\small
\begin{tabular}{@{}lccc@{}}
\toprule
           & \textbf{\# images} & \textbf{\# captions} & \textbf{vocab. size} \\ \midrule
COCO train & 113,287    & 566,435      & 9,486        \\
$COCO_{text}$ & 113,287    & 2,423,730     & 19,952       \\
$COCO_{image}$  & 113,287    & 566,435      & 9,486        \\
$SD_{base}$   & 566,747    & 566,747      & 9,486        \\
$SD_{true}$   & 566,747    & 2,835,615     & 9,486        \\
$SD_{para}$   & 566,747    & 2,423,730     & 19,952       \\ \bottomrule
\end{tabular}
\caption{Descriptive statistics of the COCO training set and its augmented datasets. }
\label{data-stats}
\end{table}

Since $COCO_{text}$ does not change images of the original COCO dataset and $COCO_{image}$ replaces all COCO images with transformed images, the number of images in both remains the same. The images in three augmented $SD$ datasets are generated based on COCO captions, so the number of images is the same as the number of COCO captions. Textual augmentations are applied to $COCO_{text}$ and $SD_{para}$. So the number of captions increases. Meanwhile, novel words are introduced by paraphrasing models, so the vocabulary sizes of the two datasets are also expanded. Table \ref{para-ex} lists a few examples of paraphrased captions. 


\begin{table}[h]
\centering
\small
\begin{tabular}{@{}c@{}}
\toprule
\textbf{COCO captions} \quad  $\rightarrow$ \quad \textbf{paraphrases} \\ \midrule
\makecell[l]{A young boy standing in front of a computer keyboard. \\→ A young boy standing before a computer keyboard.}\\ \midrule
\makecell[l]{A man is in a kitchen making pizzas. \\→ In a kitchen, a man is making pizzas.}\\ \midrule
\makecell[l]{A woman eating vegetables in front of a stove. \\→ A woman consuming vegetables in front of a cooker.}\\ \midrule
\makecell[l]{A toilet and a sink in small bathroom. \\→ A bathroom with a toilet and a sink.}\\ \midrule
\makecell[l]{A city street filled with traffic and parking lights. \\→ A city street crammed with traffic and parking lights.}\\ \bottomrule
\end{tabular}
\caption{Examples of COCO captions and their corresponding paraphrases}
\label{para-ex}
\end{table}

For synthetic image generation, we did not intentionally tune the prompts for \textit{Stable Diffusion} to obtain higher quality. Because the prompt tuning is too time-consuming when generating such a huge synthetic dataset. And image aesthetics is not our focus. Therefore, the COCO captions are directly used as input without any modifications. 


Most generated images show clear objects that are recognizable to humans (see the top two rows in Figure \ref{img-ex}), while others suffer from problems due to the limitations of text-to-image models. For example, \textit{Stable Diffusion} is not good at drawing human hands and faces, and weird distortions of objects may occur (see the bottom two rows in Figure \ref{img-ex}). Nonetheless, trained on these imperfect synthetic images, we have already shown that image captioning models can achieve quite good performance. 

Since the \textit{Stable Diffusion} model used to generate images and the T5 model for paraphrasing are open-source models, there are no issues concerning copyrights of the generated images and textual data. We also open access to our synthetic COCO dataset which can be used freely for further research.

\section{Experimental Results}

Here we list the detailed results of all three experiments. 

Table \ref{exp1-full} shows the results of both the FC model and Transformer-based model in Experiment 1, in which we trained captioning models with our three synthetic $SD$ datasets. 


In Experiment 2, we perform data augmentation on limited COCO data, mixing sampled COCO dataset with our synthetic $SD$ datasets. The baseline models are trained on these augmented datasets, and we compare the image captioning performance between our proposed method and the baseline data augmentation methods. The results of the FC model and Transformer-based model are shown in Table \ref{exp2-fc-full} and Table \ref{exp2-tfm-full}, respectively. The performances of the COCO dataset combined with the selected synthetic datasets are also listed in Table \ref{exp2-fc-full} and Table \ref{exp2-tfm-full} for convenient comparison.

We checked the relationship between synthetic data quality and their downstream performance in Experiment 3. The captioning performances of the two baseline models trained on selected $SD_{true}$ datasets are presented in Table \ref{exp3-full-select}, compared with $SD_{true}$ and $SD_{para}$. The results of data quality selection in scarce-data situations are also shown in Table \ref{exp2-fc-full} and Table \ref{exp2-tfm-full}.

\begin{table}[h]
\centering
\small
\begin{tabular}{|c|c|}
\hline
\textbf{COCO images} & \textbf{Generated images} \\ \hline
\makecell[c]{
\includegraphics[width=0.2\textwidth]{img/365525_0.jpg}
} &  \makecell[c]{\\
\includegraphics[width=0.2\textwidth]{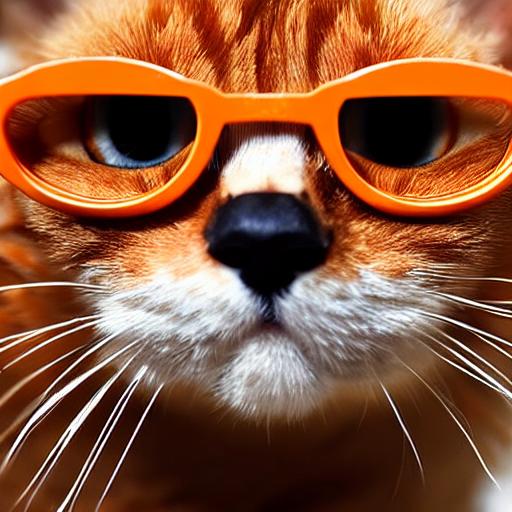}
\\ \quad}         \\ \hline
\makecell[c]{\\
\includegraphics[width=0.2\textwidth]{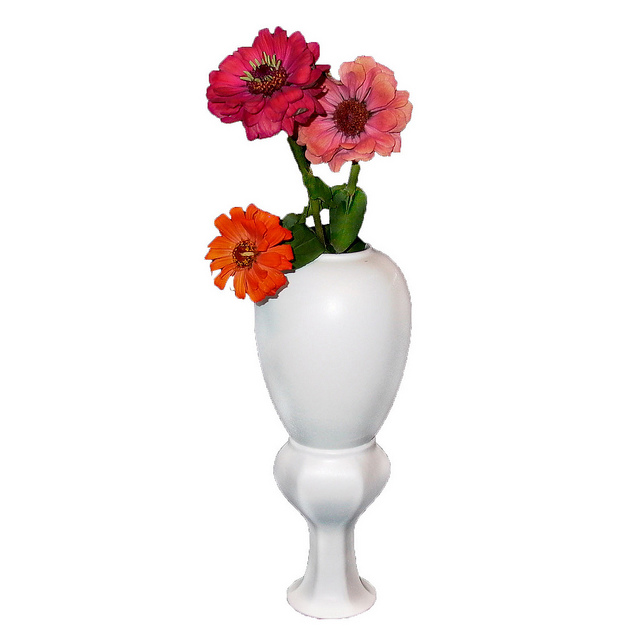}
\\ \quad}       &   \makecell[c]{\\
\includegraphics[width=0.2\textwidth]{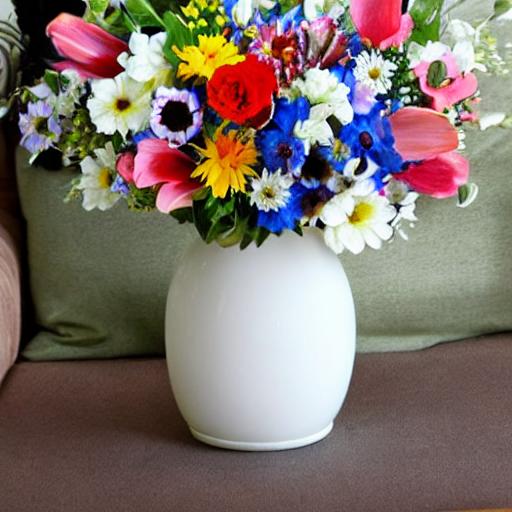}
\\ \quad}       \\ \hline
\makecell[c]{\\
\includegraphics[width=0.2\textwidth]{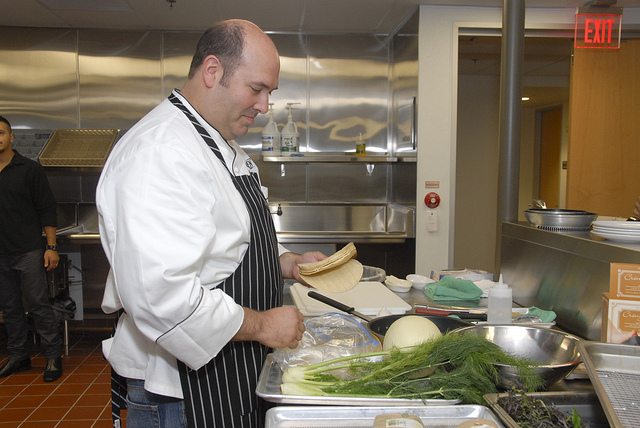}
\\ \quad}       &  \makecell[c]{\\
\includegraphics[width=0.2\textwidth]{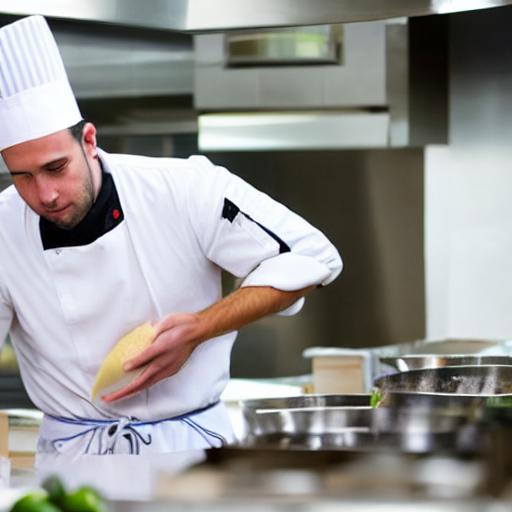}
\\ \quad}        \\ \hline
\makecell[c]{\\
\includegraphics[width=0.2\textwidth]{img/775795_0.jpg}
\\ \quad}       &  \makecell[c]{\\
\includegraphics[width=0.2\textwidth]{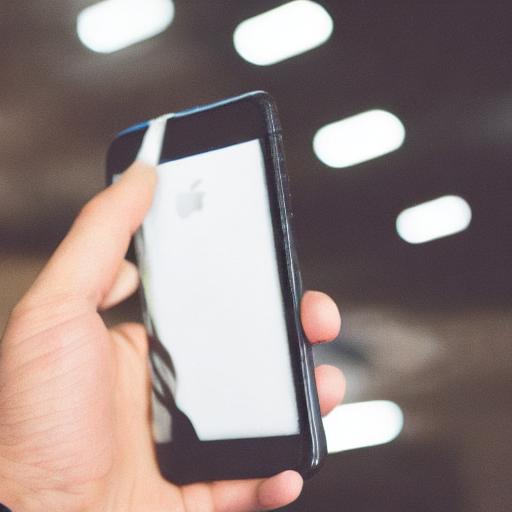}
\\ \quad}        \\ \hline
\end{tabular}
\caption{Examples of COCO and synthetic images. The COCO images are in the left column, and the right column is the corresponding generated image. }
\label{img-ex}
\end{table}

\section{Captioning Examples}

To give readers a more intuitive understanding of the improvement after applying the proposed multimodal data augmentation, we show some captioning examples in this section. The example images are sampled from the COCO test set, and the corresponding captions generated by trained models are listed in the right column.

\begin{table*}[h]
\centering
\small
\begin{tabular}{@{}l|l|cccccccc@{}}
\toprule
                         \textbf{Model} & \textbf{Training Set} &  \textbf{BLEU-1} &   \textbf{BLEU-2} &   \textbf{BLEU-3} & \textbf{BLEU-4} & \textbf{METEOR} & \textbf{ROUGE} & \textbf{CIDEr} & \textbf{SPICE} \\ \midrule
\multirow{4}{*}{FC model} & COCO & 72.2 & 54.6 & 39.7 & 28.7   & 24.4   & 52.4  & 92.4  & 17.6  \\
                          & $SD_{base}$  & 68.4 & 49.3 & 34.2      & 23.6   & 21.7   & 48.6  & 75.4  & 15.3  \\
                          & $SD_{para}$  & 69.2 & 50.1 & 34.5    & 23.4   & 21.6   & 48.9  & 75.5  & 15.4  \\ 
                          & $SD_{true}$  & \textbf{69.5} & \textbf{50.9} & \textbf{36.1}      & \textbf{25.6}   & \textbf{22.6}   & \textbf{50.1}  & \textbf{81.0}    & \textbf{15.6}  \\ \midrule
\multirow{4}{*}{T model}  & COCO & 75.5 & 59.2 & 44.7 & 33.5   & 27.4   & 55.8  & 111.3 & 20.6  \\
                          & $SD_{base}$  & 70.0 & 52.0 & 37.2      & 26.2   & 23.8   & 50.6  & 87.5  & 17.5  \\
                          & $SD_{para}$ & \textbf{71.5} & \textbf{53.5} & 38.1       & 26.8   & 23.9   & 51.2  & 89.0  & 18.0  \\
                          & $SD_{true}$ & 70.4 & 53.1 & \textbf{38.9}       &   \textbf{28.4} & \textbf{25.1}  & \textbf{52.2}  &  \textbf{94.1}  & \textbf{17.8}      \\ \bottomrule
\end{tabular}
\caption{Captioning performance of the FC model and Transformer-based model trained on the three synthetic datasets. }
\label{exp1-full}
\end{table*}

\begin{table*}[h]
\centering
\small
\begin{tabular}{@{}l|cccccccc@{}}
\toprule
              \textbf{Training Set}      & \textbf{BLEU-1} & \textbf{BLEU-2} & \textbf{BLEU-3} & \textbf{BLEU-4} & \textbf{METEOR} & \textbf{ROUGE} & \textbf{CIDEr} & \textbf{SPICE} \\ \midrule
10\% COCO           & 68.0   & 49.4   & 34.6   & 24.3   & 22.0   & 49.1  & 75.5  & 15.0  \\
10\% $COCO_{text}$    & 68.0   & 49.6   & 34.5   & 23.8   & 21.6   & 49.1  & 74.6  & 15.0  \\
10\% $COCO_{img}$     & 67.5   & 49.1   & 34.4   & 24.1   & 22.0   & 49.0  & 75.7  & 15.0  \\
10\% COCO + $SD_{base}$ & 69.9   & 51.8   & 36.8   & 25.8   & 23.1   & 50.3  & 83.5  & 16.4  \\
10\% COCO + $SD_{para}$ & 70.5   & 52.3   & 37.0   & 26.0   & 22.9   & 50.4  & 84.2  & \textbf{16.6}  \\
10\% COCO + $SD_{true}$ & 70.0   & 51.9   & 36.9   & 26.2   & 23.1   & 50.7  & 84.7  & 16.1  \\
10\% COCO + selected $SD_{true}$ & \textbf{70.6}   & \textbf{52.5}   & \textbf{37.6}   & \textbf{26.8}   & \textbf{23.3}   & \textbf{51.1}  & \textbf{86.2}  & 16.5  \\ \midrule
20\% COCO           & 69.2   & 50.9   & 36.3   & 25.7   & 22.8   & 50.2  & 80.6  & 15.8  \\
20\% $COCO_{text}$    & 69.2   & 51.0   & 35.9   & 24.9   & 22.2   & 49.9  & 79.1  & 15.7  \\
20\% $COCO_{img}$     & 69.4   & 51.5   & 36.9   & 26.3   & 22.9   & 50.4  & 82.5  & 16.0  \\
20\% COCO + $SD_{base}$ & 70.7   & 52.5   & 37.4   & 26.5   & 23.4   & 50.8  & 85.5  & 16.8  \\
20\% COCO + $SD_{para}$ & \textbf{71.5}   & \textbf{53.0}   & 37.5   & 26.3   & 23.1   & 50.8  & 86.1  & 16.8  \\
20\% COCO + $SD_{true}$ &  70.6   &  52.4  &  37.6   &  26.9  &  23.4  &  51.0   &  86.9  &  16.6  \\
20\% COCO + selected $SD_{true}$ &  70.7   &  52.6  &  \textbf{37.8}   &  \textbf{27.1}  &  \textbf{23.5}  &  \textbf{51.2}   &  \textbf{87.2}  &  \textbf{17.0}  \\ \midrule
30\% COCO           & 70.2   & 52.2   & 37.4   & 26.6   & 23.3   & 50.9  & 85.1  & 16.4  \\
30\% $COCO_{text}$    & 70.3   & 52.4   & 37.2   & 26.1   & 22.8   & 50.9  & 83.1  & 16.1  \\
30\% $COCO_{img}$     & 69.9   & 52.1   & 37.4   & 26.7   & 23.4   & 50.9  & 85.4  & 16.3  \\
30\% COCO + $SD_{base}$      & 70.5   & 52.3   & 37.3   & 26.3   & 23.4   & 50.8  & 85.5  & 16.6  \\
30\% COCO + $SD_{para}$ & \textbf{71.6}   & \textbf{53.5}   & \textbf{38.3}   & 27.0   & 23.5   & 51.4  & 87.2  & \textbf{17.0}  \\
30\% COCO + $SD_{true}$ &  70.4  &  52.4  &  37.6  &  26.9  &  23.4  & 51.2  &  86.0  &  16.6  \\
30\% COCO + selected $SD_{true}$ &  70.9  &  52.9  &  38.0  &  \textbf{27.0}  &  \textbf{23.7}  & \textbf{51.4}  &  \textbf{87.8}  &  16.9  \\ \midrule
40\% COCO           & 71.0   & 53.1   & 38.3   & 27.4   & 23.6   & 51.5  & 87.8  & 16.7  \\
40\% $COCO_{text}$    & 70.3   & 52.4   & 37.2   & 26.2   & 22.7   & 51.0  & 83.6  & 16.1  \\
40\% $COCO_{img}$     & 70.7   & 52.9   & 38.0   & 27.2   & 23.6   & 51.3  & 87.5  & 16.6  \\
40\% COCO + $SD_{base}$      & 70.8   & 52.7   & 37.6   & 26.6   & 23.5   & 51.1  & 86.0  & 16.7  \\
40\% COCO + $SD_{para}$ & \textbf{71.6}   & \textbf{53.7}   & \textbf{38.5}   & 27.3   & 23.6   & \textbf{51.4}  & \textbf{88.5}  & \textbf{17.0}  \\
40\% COCO + $SD_{true}$ &  70.4   & 52.7  &  37.8 &   27.1  &  23.6  &  51.3  &  87.1  &  16.6  \\
40\% COCO + selected $SD_{true}$ &  70.9   & 52.9  &  38.1 &   \textbf{27.3}  &  \textbf{23.7}  &  51.4  &  88.3  &  17.0  \\ \midrule
50\% COCO           & 71.3   & 53.5   & 38.5   & 27.5   & 23.7   & 51.5  & 88.1  & 16.8  \\
50\% $COCO_{text}$    & 70.4   & 52.6   & 37.4   & 26.4   & 22.9   & 51.1  & 84.3  & 16.3  \\
50\% $COCO_{img}$     & 71.1   & 53.3   & 38.5   & 27.7   & 23.8   & 51.8  & 88.3  & 16.8  \\
50\% COCO + $SD_{base}$      & 70.8   & 52.7   & 37.7   & 26.8   & 23.6   & 51.2  & 87.3  & 16.8  \\
50\% COCO + $SD_{para}$ & \textbf{71.8}   & \textbf{53.9}   & \textbf{38.7}   & 27.5   & 23.7   & \textbf{51.6}  & 88.6  & \textbf{17.2}  \\
50\% COCO + $SD_{true}$ &  70.8  & 52.8   &  37.9  & 27.1  &  23.6  & 51.3  & 87.3  &  16.8  \\
50\% COCO + selected $SD_{true}$ &  71.2  & 53.4   &  38.5  & \textbf{27.6}  &  \textbf{23.7}  & 51.5  & \textbf{89.3}  &  16.9  \\ \bottomrule
\end{tabular}

\caption{Data augmentation performance of FC model with limited true COCO data, including the performance of data quality selection in the last row of each group. }
\label{exp2-fc-full}
\end{table*}

\begin{table*}[h]
\centering
\small
\begin{tabular}{@{}l|cccccccc@{}}
\toprule
                  \textbf{Training Set}  & \textbf{BLEU-1} & \textbf{BLEU-2} & \textbf{BLEU-3} & \textbf{BLEU-4} & \textbf{METEOR} & \textbf{ROUGE} & \textbf{CIDEr} & \textbf{SPICE} \\ \midrule
10\% COCO           & 69.3   & 51.5   & 37.1   & 26.3   & 24.1   & 50.8  & 88.2  & 17.6  \\
10\% $COCO_{text}$    & 69.5   & 51.9   & 36.9   & 25.9   & 23.9   & 50.5  & 87.1  & 17.3  \\
10\% $COCO_{img}$     & 68.7   & 50.8   & 36.3   & 25.7   & 23.6   & 50.5  & 85.3  & 17.0  \\
10\% COCO + $SD_{base}$ & 72.3   & 54.8   & 39.8   & 28.7   & 24.9   & 52.1  & 95.6  & 18.7  \\
10\% COCO + $SD_{para}$ & \textbf{72.3}   & 54.5   & 39.2   & 27.7   & 24.7   & 51.8  & 94.1  & \textbf{18.8}  \\
10\% COCO + $SD_{true}$ &  71.5  &  54.5  &  40.1  &  29.4  &  \textbf{25.5}   &  53.0  & 97.5  &  18.6  \\
10\% COCO + selected $SD_{true}$ &  72.1  &  \textbf{54.9}  &  \textbf{40.5}  &  \textbf{29.5}  &  25.4   &  \textbf{53.1}  & \textbf{98.3}  &  18.7  \\ \midrule
20\% COCO           & 71.0   & 53.5   & 39.1   & 28.2   & 25.3   & 52.3  & 94.8  & 18.9  \\
20\% $COCO_{text}$    & 71.7   & 53.9   & 38.5   & 27.0   & 24.6   & 52.2  & 92.7  & 18.5  \\
20\% $COCO_{img}$     & 70.6   & 53.3   & 38.9   & 28.2   & 25.2   & 52.1  & 94.2  & 18.5  \\
20\% COCO + $SD_{base}$ & 72.5   & 55.1   & 40.1   & 28.8   & 25.7   & 52.8  & 99.2  & 19.4  \\
20\% COCO + $SD_{para}$ & \textbf{74.0}   & \textbf{56.6}   & 41.5   & 30.1   & 25.7   & 53.3  & 100.5 & 19.2  \\
20\% COCO + $SD_{true}$ & 73.2 &  56.5  &  \textbf{42.1}  &  \textbf{31.1}  & \textbf{26.2}   & \textbf{53.9}  & \textbf{103.4}  &  \textbf{19.4}  \\
20\% COCO + selected $SD_{true}$ & 72.0 &  55.0  &  40.6  &  29.7  & 25.8   & 53.1  & 99.9  &  19.2  \\ \midrule
30\% COCO           & 72.5   & 55.3   & 40.8   & 29.8   & 25.7   & 53.2  & 99.7  & 19.4  \\
30\% $COCO_{text}$    & 73.5   & 56.4   & 41.5   & 29.8   & 25.1   & 53.3  & 97.3  & 19.0  \\
30\% $COCO_{img}$     & 71.3   & 54.0   & 39.7   & 29.0   & 25.8   & 52.7  & 98.2  & 19.0  \\
30\% COCO + $SD_{base}$   & 73.5   & 56.4   & 41.7   & 30.6   & 25.9   & 53.6  & 102.4 & 19.5  \\
30\% COCO + $SD_{para}$ & \textbf{74.4}   & \textbf{57.6}   & \textbf{42.8}   & 31.2   & 26.1   & 54.1  & 104.2 & 19.5  \\
30\% COCO + $SD_{true}$ & 73.8  &  57.0  & 42.6 &  \textbf{31.4}  & \textbf{26.5}  & \textbf{54.6}  & \textbf{104.6}  & \textbf{19.8}  \\
30\% COCO + selected $SD_{true}$ & 72.4  &  55.5  & 41.2 &  30.4  & 26.1  & 53.6  & 102.2  & 19.2  \\ \midrule
40\% COCO           & 73.4   & 56.4   & 42.2   & 31.2   & 26.5   & 54.2  & 103.9 & 20.0  \\
40\% $COCO_{text}$    & 73.3   & 55.8   & 40.6   & 28.8   & 25.2   & 53.0  & 97.0  & 19.2  \\
40\% $COCO_{img}$     & 73.0   & 55.8   & 41.3   & 30.4   & 26.2   & 53.8  & 101.5 & 19.6  \\
40\% COCO + $SD_{base}$   & 74.1   & 57.0   & 42.4   & 31.1   & 26.3   & 54.2  & 104.5 & 19.9  \\
40\% COCO + $SD_{para}$ & \textbf{74.9}   & \textbf{57.9}   & 43.0   & 31.7   & 26.5   & 54.5  & 106.5 & \textbf{20.0}  \\
40\% COCO + $SD_{true}$ &  74.1  &  57.6  & \textbf{43.3}  &  \textbf{32.2}  & \textbf{26.9}  & \textbf{55.0}  & \textbf{106.9} & 19.9  \\
40\% COCO + selected $SD_{true}$ &  72.8  &  56.1  & 41.8  &  31.0  & 26.2  & 54.1  & 103.1 & 19.4  \\ \midrule
50\% COCO           & 74.3   & 57.3   & 42.5   & 31.0   & 26.7   & 54.6  & 105.1 & 20.0  \\
50\% $COCO_{text}$    & 74.4   & 57.6   & 42.6   & 31.1   & 25.7   & 54.1  & 102.3 & 19.5  \\
50\% $COCO_{img}$     & 73.6   & 56.5   & 42.1   & 31.1   & 26.5   & 54.1  & 103.3 & 19.7  \\
50\% COCO + $SD_{base}$  & 73.8   & 57.0   & 42.5   & 31.5   & 26.4   & 54.2  & 105.5 & 19.8  \\
50\% COCO + $SD_{para}$ & \textbf{74.9}   & \textbf{58.3}   & \textbf{43.7}   & \textbf{32.6}   & \textbf{26.8}   & \textbf{55.0}  & \textbf{107.8} & \textbf{20.1}  \\
50\% COCO + $SD_{true}$ & 74.6  & 58.0  &  43.4 &   32.2 & 26.7  & 54.8 & 107.1  &  19.9 \\
50\% COCO + selected $SD_{true}$ & 73.2  & 56.5  &  42.3 &   31.2 & 26.7  & 54.4 & 104.7  &  19.8 \\ \bottomrule
\end{tabular}
\caption{Data augmentation performance of Transformer-based model with limited true COCO data, including the performance of data quality selection in the last row of each group. }
\label{exp2-tfm-full}
\end{table*}

\begin{table*}[h]
\centering
\small
\begin{tabular}{@{}l|l|cccccccc@{}}
\toprule
                         \textbf{Model} & \textbf{Training Set}    & \textbf{BLEU-1} & \textbf{BLEU-2} & \textbf{BLEU-3} & \textbf{BLEU-4} & \textbf{METEOR} & \textbf{ROUGE} & \textbf{CIDEr} & \textbf{SPICE} \\ \midrule
\multirow{7}{*}{FC Model} & COCO                 & 72.2   & 54.6   & 39.7   & 28.7   & 24.4   & 52.4  & 92.4  & 17.6  \\ \cmidrule(l){2-10} 
                          & $SD_{base}$          & \textbf{68.4}   & \textbf{49.3}   & \textbf{34.2}   & \textbf{23.6}   & 21.7   & 48.6  & \textbf{75.4}  & 15.3  \\
                          & Selected $SD_{base}$ & 67.4   & 48.8   & 34.0   & 23.5   & \textbf{22.2}   & \textbf{48.9}  & 75.3  & \textbf{15.6}  \\ \cmidrule(l){2-10} 
                          & $SD_{para}$          & \textbf{69.2}   & \textbf{50.1}   & 34.5   & \textbf{23.4}   & 21.6   & \textbf{48.9}  & \textbf{75.5}  & 15.4  \\
                          & Selected $SD_{para}$ & 67.5   & 48.8   & \textbf{33.8}   & 23.2   & \textbf{22.0}   & 48.7  & 75.0  & \textbf{15.6}  \\ \cmidrule(l){2-10} 
                          & $SD_{true}$          & 69.5   & 50.9   & 36.1   & \textbf{25.6}   & 22.6   & 50.1  & 81.0  & 15.6  \\
                          & Selected $SD_{true}$ & \textbf{69.6}   & \textbf{51.3}   & \textbf{36.3}   & 25.5   & \textbf{22.7}   & \textbf{50.2}  & \textbf{81.0}  & \textbf{15.7}  \\ \midrule
\multirow{7}{*}{T Model}  & COCO                 & 75.5   & 59.2   & 44.7   & 33.5   & 27.4   & 55.8  & 111.3 & 20.6  \\ \cmidrule(l){2-10} 
                          & $SD_{base}$          & \textbf{70.0}   & \textbf{52.0}   & 37.2   & \textbf{26.2}   & 23.8   & 50.6  & 87.5  & 17.5  \\
                          & Selected $SD_{base}$ & 69.9   & 52.0   & \textbf{37.2}   & 26.1   & \textbf{24.1}   & \textbf{50.7}  & \textbf{87.9}  & \textbf{17.9}  \\ \cmidrule(l){2-10} 
                          & $SD_{para}$          & \textbf{71.5}   & \textbf{53.5}   & \textbf{38.1}   & \textbf{26.8}   & 23.9   & \textbf{51.2}  & 89.0  & \textbf{18.0}  \\
                          & Selected $SD_{para}$ & 70.9   & 53.0   & 37.9   & 26.7   & \textbf{24.0}   & 51.1  & \textbf{89.4}  & 17.9  \\ \cmidrule(l){2-10} 
                          & $SD_{true}$          & \textbf{70.4}   & \textbf{53.1}   & \textbf{38.9}   & \textbf{28.4}   & \textbf{25.1}   & \textbf{52.2}  & \textbf{94.1}  & 17.8  \\
                          & Selected $SD_{true}$ & 70.3   & 52.9   & 38.5   & 27.7   & 24.7   & 51.8  & 93.0  & \textbf{17.8}  \\ \bottomrule
\end{tabular}
\caption{Image captioning performance of FC model and Transformer-based model trained on fully synthetic datasets after selecting high-quality data based on the top $50\%$ \textit{CLIPScore} criterion. }
\label{exp3-full-select}
\end{table*}

\begin{table*}[h]
\centering
\small
\begin{tabular}{|c|c|}
\hline
\textbf{Images} & \textbf{Captions} \\ \hline

\makecell[c]{\\
\begin{minipage}[b]{0.3\textwidth}
    \centering
    \raisebox{-.5\height}{\includegraphics[height=\linewidth]{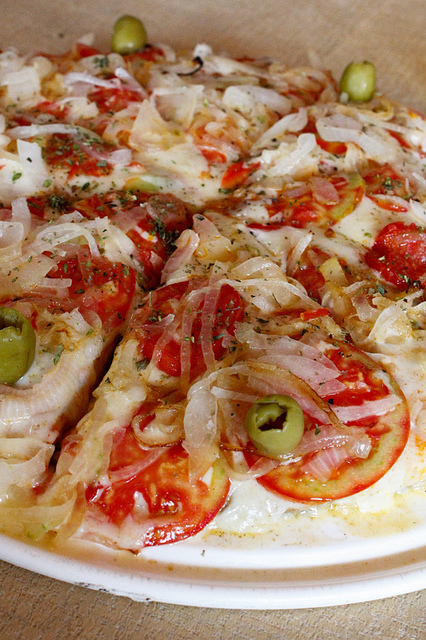} }
\end{minipage}
\\ \quad}&   \makecell[l]{\textbf{Human}: Baked pizza with red tomatoes and green olives. \\ \textbf{FC models}: \\ \textbf{10\% COCO}: a pizza \wrong{with a pizza} on a white plate \\ \textbf{10\% $\mathbf{COCO_{text}}$}: a pizza \key{with cheese and tomatoes} on a plate  \\ \textbf{10\% $\mathbf{COCO_{img}}$}: a pizza \key{with cheese and tomatoes} on a table \\ \textbf{10\% COCO + $\mathbf{SD_{base}}$}: a \key{close up} of a pizza with a lot of \key{toppings} \\ \textbf{10\% COCO + $\mathbf{SD_{para}}$}: a pizza \key{with cheese and tomatoes} on a white plate \\ \textbf{10\% COCO + $\mathbf{SD_{true}}$}: a pizza \wrong{with a slice of pizza} on it }        \\ \hline

\makecell[c]{\\
\begin{minipage}[b]{0.3\textwidth}
    \centering
    \raisebox{-.5\height}{\includegraphics[width=\linewidth]{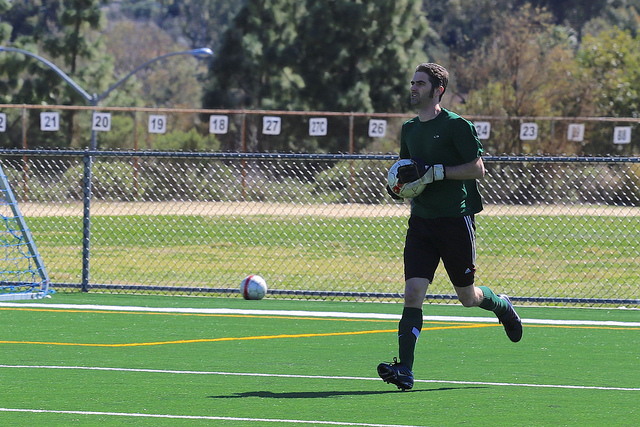} }
\end{minipage}
\\ \quad}&   \makecell[l]{\textbf{Human}: The soccer player is bringing back the ball into play. \\ \textbf{FC models}: \\ \textbf{10\% COCO}: a man is playing a game of \wrong{baseball} \\ \textbf{10\% $\mathbf{COCO_{text}}$}: a man in a \wrong{blue} shirt and \wrong{white} shorts playing a game of \wrong{baseball}  \\ \textbf{10\% $\mathbf{COCO_{img}}$}: a man is playing a game of \key{soccer} on a field \\ \textbf{10\% COCO + $\mathbf{SD_{base}}$}: a man is playing a game of \key{soccer} \\ \textbf{10\% COCO + $\mathbf{SD_{para}}$}: a man in a green shirt and a \wrong{baseball} uniform \\ \textbf{10\% COCO + $\mathbf{SD_{true}}$}: a man in a \wrong{red} shirt is playing \key{soccer} }        \\ \hline

\makecell[c]{\\
\begin{minipage}[b]{0.3\textwidth}
    \centering
    \raisebox{-.5\height}{\includegraphics[width=\linewidth]{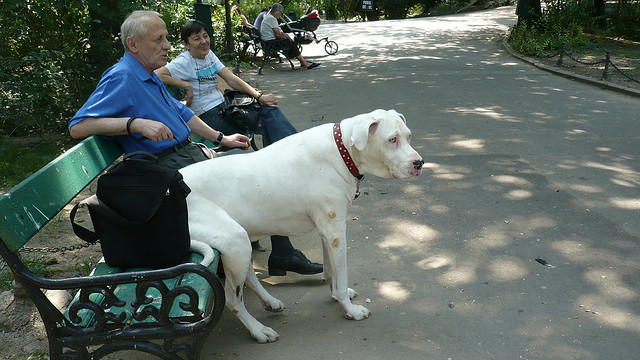} }
\end{minipage}
\\ \quad}&   \makecell[l]{\textbf{Human}: A large white dog is sitting on a bench beside an elderly man. \\ \textbf{FC models}: \\ \textbf{10\% COCO}: a dog and \wrong{a dog} are on a bench \\ \textbf{10\% $\mathbf{COCO_{text}}$}: \wrong{a couple of dogs} sitting on a bench  \\ \textbf{10\% $\mathbf{COCO_{img}}$}: a dog and \wrong{a dog} are sitting on a bench \\ \textbf{10\% COCO + $\mathbf{SD_{base}}$}: a dog is sitting on a bench \wrong{with a dog} \\ \textbf{10\% COCO + $\mathbf{SD_{para}}$}: \key{a man} sitting on a bench with a dog \\ \textbf{10\% COCO + $\mathbf{SD_{true}}$}: \key{a man} sitting on a bench with a dog }        \\ \hline

\makecell[c]{\\
\begin{minipage}[b]{0.3\textwidth}
    \centering
    \raisebox{-.5\height}{\includegraphics[height=\linewidth]{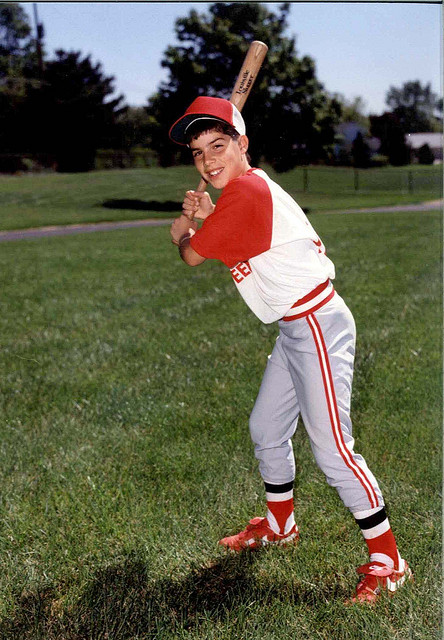} }
\end{minipage}
\\ \quad}&   \makecell[l]{\textbf{Human}: A little boy in a baseball uniform holds the bat ready to swing. \\ \textbf{Transformer-based models}: \\ \textbf{COCO}: \key{a young man} holding a \key{baseball bat} in a park \\ \textbf{$\mathbf{SD_{base}}$}: a man in a \key{baseball uniform swinging a bat} \\ \textbf{$\mathbf{SD_{para}}$}: \key{a young man} holding \key{a baseball bat} on a field \\ \textbf{$\mathbf{SD_{true}}$}: \key{a young man} holding \key{a baseball bat} on a field }        \\ \hline
\end{tabular}
\caption{Captions generated by baseline models trained on the COCO dataset and our synthetic datasets. \wrong{Red} indicates wrong objects detected or poor use of words (e.g., repeating existing words); \key{Blue} highlights correct objects or good use of words. We can observe that our method outperforms the baselines. And the quality of captions generated by our method is close to human-written sentences. }
\label{cap-ex}
\end{table*}

\end{document}